\documentclass[11pt]{article}

\usepackage{acl}
\usepackage{times}
\usepackage{latexsym}
\usepackage[T1]{fontenc}
\usepackage[utf8]{inputenc}
\usepackage{microtype}

\usepackage{amsmath, amssymb, amsthm}
\usepackage{graphicx}
\usepackage{booktabs}
\usepackage{multirow}
\usepackage{xcolor}
\usepackage{hyperref}
\usepackage{cleveref}
\usepackage{algorithm}
\usepackage{algorithmicx}
\usepackage{algpseudocode}
\usepackage{afterpage}
\usepackage{float}
\newcommand{\error}[1]{\textcolor{red!75!black}{\textbf{\texttt{#1}}}}

\definecolor{acceptcol}{RGB}{30,130,40}
\definecolor{fusecol}{RGB}{210,110,30}
\newcommand{\accepted}[1]{\textcolor{acceptcol}{\textbf{#1}}}
\newcommand{\fused}[1]{\textcolor{fusecol}{\textbf{#1}}}
\newcommand{\crchange}[1]{#1}

\theoremstyle{plain}

\theoremstyle{definition}

\title{AsymSpec: Context-Asymmetric Speculative Decoding for Agentic LLMs}

\author{
 \textbf{Sheng Liang\textsuperscript{1}}, 
 \textbf{Yongyue Zhang\textsuperscript{1}},
 \textbf{Nathanael Brian\textsuperscript{1}},
 \textbf{Hang Lv\textsuperscript{2}},\\
 \textbf{Hao Wang\textsuperscript{2}},
 \textbf{Chen Zhang\textsuperscript{1}},
 \textbf{Yong Liu\textsuperscript{1}},
 \\
\textsuperscript{1}Huawei Technologies Co., Ltd. 
\\ 
\textsuperscript{2}University of Science and Technology of China
\\
}

\begin{document}
\maketitle

\begin{abstract}
Agentic LLM pipelines face escalating inference costs as context accumulates across retrieval, tool use, and multi-turn interactions. To control latency, deployments routinely compress inputs, but this degrades task accuracy. Speculative decoding (SD) accelerates generation losslessly, yet it assumes the drafter and verifier share an identical context, preventing SD from resolving the accuracy--overhead trade-off. We propose \textsc{AsymSpec}, an asymmetric speculative decoding framework that breaks this symmetry: a lightweight drafter reads the full input while the large verifier operates on the compressed view. The drafter steers the verifier via a contrastive $\delta$-fusion of logits, modulated by a divergence-aware acceptance gate that preserves verification stability and high draft acceptance rates. Evaluated across four agentic capabilities and two end-to-end agent benchmarks, \textsc{AsymSpec} reaches $\approx$90\% of full-context accuracy on average, delivering 1.3--1.7$\times$ throughput speedups at 0.2--0.3$\times$ the compute cost on isolated text capabilities. These results show that asymmetric context access yields substantial gains precisely when compression discards critical reasoning signals.
\end{abstract}

\section{Introduction}
\label{sec:intro}

Modern LLM deployments increasingly operate as agentic pipelines---retrieving documents~\crchange{\citep{lewis2020rag,gao2023ragsurvey,wu-etal-2025-querycentric}}, invoking tools~\citep{yao2023react,toolformer2023}, maintaining multi-turn dialogue\crchange{ and memory}~\crchange{\citep{sirdeshmukh2025multichallenge,DBLP:journals/corr/abs-2501-09959,wu-etal-2025-memory}}, and processing multimodal inputs~\citep{liu2024llava}. These settings issue repeated LLM calls over context that grows with every step. As retrieved passages, tool observations, and interaction histories accumulate, the forward pass becomes the dominant latency bottleneck, making context length the primary driver of inference overhead in production.

To control this overhead, deployments routinely \emph{compress} the context~\citep{llmlingua2024}: RAG pipelines summarize retrieved passages, tool-use agents pass only API signatures instead of full documentation, and multimodal workflows feed short captions instead of raw images. Compression reduces serving cost but systematically discards fine-grained details critical for task accuracy. Deployments thus face a rigid accuracy--overhead trade-off: absorb the prohibitive latency of full-context generation, or accept significant degradation.

Speculative decoding (SD)~\citep{leviathan2023fast,chen2023accelerating} has become the standard approach for accelerating LLM inference: a lightweight drafter proposes candidate tokens that a large verifier checks in parallel, guaranteeing lossless generation under the target distribution. Despite extensive improvements to drafter architectures~\citep{li2024eagle,cai2024medusa,li2024eagle2} and contrastive logit fusion in the speculative loop~\citep{yuan2023scd}, all existing SD methods share a foundational constraint: drafter and verifier process the same input tokens. SD accelerates a fixed target distribution without changing what the target model sees; once the verifier is compressed, SD can only accelerate the compressed model---it cannot recover what compression removed. Either both models pay the full-context cost, or both inherit the compression loss.

Our key observation is a structural \emph{compute asymmetry}: per-step latency is dominated by the large verifier, while a lightweight drafter adds negligible overhead. Compressing only the verifier captures most of the latency savings, but standard SD cannot exploit this because it enforces identical input. We break this symmetry with \textsc{AsymSpec}, an asymmetric speculative decoding framework that explicitly decouples context access. The verifier operates strictly on the compressed view for efficiency, while the drafter reads the full input to reconstruct the discarded information. We realize this recovery through a contrastive mechanism: the drafter processes both context views, and subtracting their output distributions removes the drafter's context-independent preferences, isolating the information gain the uncompressed input provides. This gain signal, $\delta$, is fused into the verifier's logits and modulated by a parameter-free Context-Divergence Acceptance (CDA) gate. By scaling injection strength with the context divergence, the gate maintains stable verification and high acceptance rates. Since only the drafter processes the full input, \textsc{AsymSpec} recovers full-context reasoning fidelity at the compressed verifier's latency, and extends to cross-modal settings (e.g., a vision--language drafter on raw images steering a text-only verifier on captions).

\paragraph{Contributions.}
\vspace{-0.2em}
\begin{enumerate}\itemsep0pt
\vspace{-0.2em}
\item We propose \textsc{AsymSpec}, a context-asymmetric speculative
  decoding framework: the verifier runs on a compressed view while the
  drafter consumes the full input, opening an operating point---compressed
  cost with near-ceiling accuracy---inaccessible to symmetric SD, and
  extending naturally to cross-modal settings.
\vspace{-0.2em}
\item Two coupled mechanisms instantiate the framework: a same-model
  cross-context $\delta$-fusion that cancels drafter capacity biases to
  isolate the context-gain signal, and a parameter-free Context-Divergence
  Acceptance (CDA) gate that bounds steering strength without per-dataset
  tuning.
\vspace{-0.2em}
\item Across four agentic capabilities and two end-to-end agent benchmarks,
  \textsc{AsymSpec} recovers $\approx$90\% of full-context accuracy at
  $0.2$--$0.3\times$ compute and $1.3$--$1.7\times$ throughput.
\end{enumerate}

\section{Related Work}
\label{sec:related}

\subsection{Speculative decoding}
Speculative decoding (SD)~\citep{leviathan2023fast,chen2023accelerating} accelerates autoregressive generation by having a lightweight drafter propose candidate tokens that a large verifier validates in parallel, preserving the target distribution via rejection sampling. Subsequent work improves drafter quality (EAGLE~\citep{li2024eagle,li2024eagle2,li2025eagle3}, Medusa~\citep{cai2024medusa}) or targets long-context inference latency via hierarchical or sparse-KV speculation (TriForce~\citep{triforce2024}, MagicDec~\citep{magicdec2025}). All these methods feed the drafter and verifier the \emph{same input tokens}, even when KV-cache structure differs across stages. This symmetry prevents SD from exploiting the compute asymmetry inherent in long-context decoding. We relax this constraint in \S\ref{sec:method}, letting the two models operate on distinct context views without breaking the speculative verification loop.

\subsection{Context compression}
Context compression is the standard mechanism for reducing long-context inference overhead~\crchange{\citep{lv-etal-2026-iecache}}. Hard-prompt methods prune tokens by importance (LLMLingua~\citep{longllmlingua2024,llmlingua2024}), soft-context approaches learn compact latents (Gist Tokens~\citep{gist2023}, ICAE~\citep{icae2024}), and KV-cache techniques operate directly on cached states (StreamingLLM~\citep{streamingllm2024}, SnapKV~\citep{snapkv2024}). Comprehensive surveys~\citep{promptcompsurvey2024} document these approaches. Across this literature, accuracy degradation is treated as an unavoidable cost. We treat any compressor as a black box and show in \S\ref{sec:method} how a full-context drafter can systematically recover the discarded information, turning compression from a lossy shortcut into a steerable efficiency knob.

\subsection{Contrastive decoding and logit fusion}
Contrastive decoding~\citep{li2023contrastive} improves generation quality by subtracting an amateur model's logits from an expert's, amplifying expert-specific signals. This principle has been extended to multi-step reasoning~\citep{obrien2023cdreasoning} and integrated into the speculative loop as Speculative Contrastive Decoding (SCD)~\citep{yuan2023scd}. However, SCD and its variants operate on a single shared context, using logit differences to bridge a \emph{model-capacity gap}. \crchange{The same subtractive principle has also been used to remove content-independent positional priors with CapCal~\citep{lv-etal-2026-learning}, mitigate linguistic inertia after reasoning-chain compression with LICD~\citep{zhang2026thinkinghurtsdiagnosingrectifying}, and transfer local context-induced preferences to a remote model with CoSteer~\citep{lv-etal-2025-costeer}. Together, these methods frame logit differences as signal isolators. \textsc{AsymSpec} brings this view into speculative verification: both logit terms come from the same drafter under full and compressed context views, so $\delta$ isolates the context gain rather than a model-capacity gap.} We formalize this cross-context transfer and its bias-cancellation property in \S\ref{sec:method-delta}.

\subsection{Asymmetric and multimodal speculation}
Recent work has begun to explore asymmetric configurations. RAPID~\citep{liu2025rapid} adopts an inverse design: a retrieval-truncated drafter steers a full-context verifier to inject external knowledge, prioritizing full-context fidelity at full-context latency. Speculative RAG~\citep{wang2024specrag} drafts answer candidates from retrieved-document subsets and then verifies them. Concurrent multimodal SD methods (SpecVLM, Spec-LLaVA, ViSpec~\citep{specvlm2025,specllava2025,vispec2025}; MASSV~\citep{massv2025}) focus on visual-token prefill acceleration or keep both models within a single modality. SD2~\citep{sd2_2025} reverses the steering direction by conditioning the drafter on verifier signals. \crchange{SpecSteer uses asymmetric local--cloud speculation, keeping private personalization context with a local drafter while drawing on cloud-scale reasoning~\citep{lv-etal-2026-specsteer}. Across these designs, asymmetry serves personalization and privacy, retrieval allocation, or modality-specific acceleration. \textsc{AsymSpec} instead uses asymmetric context access to change the efficiency operating point: a rich-input drafter compensates for a strictly compressed verifier, recovering accuracy at compressed-verifier latency} (\S\ref{sec:method}; design-space comparison in \Cref{tab:distinguish}).

\subsection{Speculation in agentic settings}
Most SD work targets single-turn generation; long-context variants~\citep{triforce2024,magicdec2025} extend the regime but remain prompt-level. Recent agent-oriented methods accelerate the decision loop by speculating over high-level action sequences~\citep{agentspec2024}, tool invocations~\citep{toolspec2024}, or plans~\citep{speculative_planning2025}. \crchange{DynaThink likewise reduces inference cost by choosing between fast and deliberative reasoning at the request level~\citep{DBLP:conf/emnlp/Pan0ZL0024}. Such action- and request-level mechanisms} are orthogonal to token-level decoding engines. We demonstrate in \S\ref{sec:results-gaia} that \textsc{AsymSpec} composes with live agentic loops, providing token-level accuracy recovery whenever tool outputs or retrieved contexts are compressed online.

\begin{figure*}[!t]
  \centering
  \includegraphics[width=0.8\textwidth]{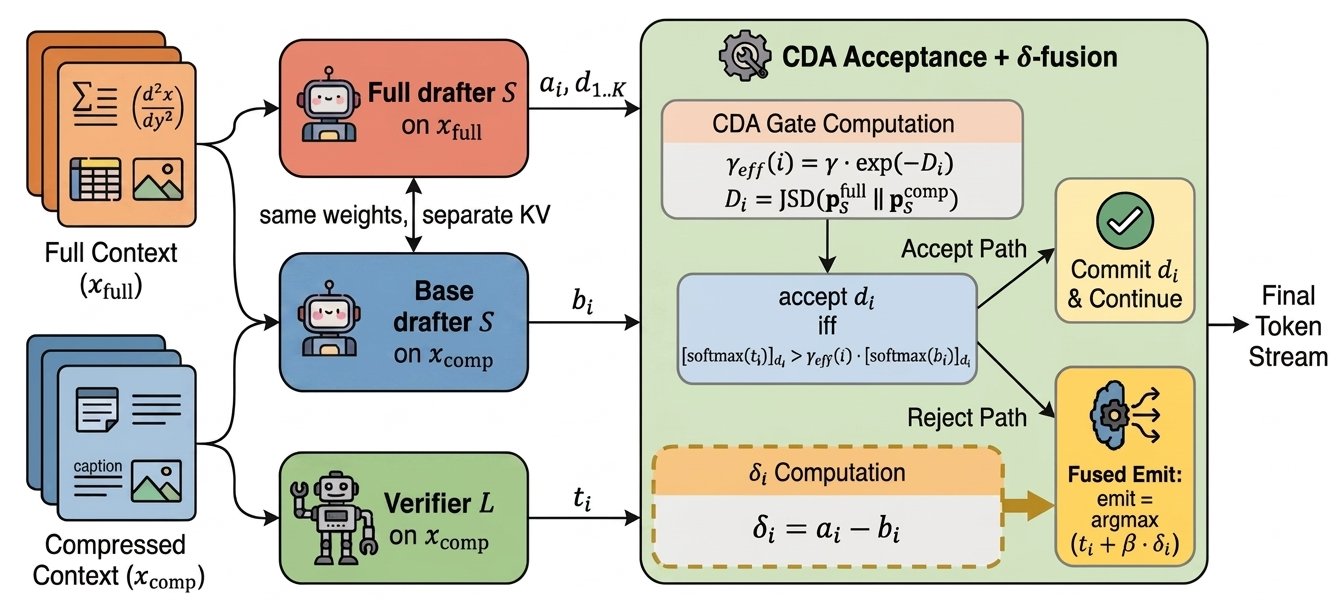}
  \caption{\textsc{AsymSpec} speculation step. \textbf{(a) Drafting \& verification:} drafter $S$ reads $x_{\text{full}}$ (logits $a$, drafts $d_{1:K}$) and $x_{\text{comp}}$ (logits $b$); verifier $L$ reads $x_{\text{comp}}$ (logits $t$). \textbf{(b) CDA gate \& $\delta$-fusion:} $\gamma_{\text{eff}}(i)$ relaxes as full-vs-compressed divergence grows; accepted drafts are committed, otherwise the $\delta$-fused $\arg\max$ is emitted. Only $S$ consumes $x_{\text{full}}$.}
  \label{fig:pipeline}
\end{figure*}

\section{Method}
\label{sec:method}

\subsection{Problem setup}
\label{sec:method-setup}
Let $L$ be a large verifier and $S$ a lightweight drafter ($|S|\ll|L|$). A task provides a full prompt $x_{\text{full}}$; a black-box compressor produces a compressed view $x_{\text{comp}}$ with $|x_{\text{comp}}|\ll|x_{\text{full}}|$. Per-step latency is dominated by $L$'s forward pass, which scales super-linearly with context length. This overhead structure defines two baseline operating points:
\begin{itemize}\itemsep0pt
\item $L(\cdot\mid x_{\text{full}})$: peak accuracy, but prohibitive latency in agentic loops;
\item $L(\cdot\mid x_{\text{comp}})$: low latency, but accuracy degraded by compression.
\end{itemize}
Compressing only $L$'s input captures the majority of latency savings, while $S$'s extra forward on $x_{\text{full}}$ remains far cheaper than the verifier's saved forward on $x_{\text{full}}$. Standard speculative decoding cannot exploit this asymmetry because it forces identical context access. \textsc{AsymSpec} explicitly decouples the two: $L$ operates strictly on $x_{\text{comp}}$ for efficiency, while $S$ reads $x_{\text{full}}$ to reconstruct discarded information.

\subsection{Contrastive $\delta$-fusion}
\label{sec:method-delta}
Each speculation step executes three forward passes over the current sequence (\Cref{fig:pipeline}a):
\begin{enumerate}\itemsep0pt
    \item \textbf{Augmented drafter:} $S(x_{\text{full}})$ produces logits $a$ and samples $K$ draft tokens $d_{1:K}$.
    \item \textbf{Base drafter:} $S(x_{\text{comp}})$ produces logits $b$ at the same $K+1$ positions.
    \item \textbf{Verifier:} $L(x_{\text{comp}})$ scores all $K$ drafts in parallel, yielding logits $t$.
\end{enumerate}
We define the context-gain signal in logit space:
\vspace{-0.8em}
\begin{equation}
\delta_i \;=\; a_i - b_i,
\label{eq:delta}
\vspace{-0.5em}
\end{equation}
where $a_i, b_i \in \mathbb{R}^{|\mathcal{V}|}$ are the per-position logits. Subtracting $b_i$ from $a_i$ removes the drafter's context-independent preferences, isolating the shift induced by the additional context. Upon draft rejection, $\delta_i$ is fused into the verifier's distribution:
\vspace{-0.8em}
\begin{equation}
d'_i \;=\; \arg\max\bigl(t_i + \beta\,\delta_i\bigr),
\label{eq:fusion}
\vspace{-0.5em}
\end{equation}
where $\beta\in[0,1]$ controls the steering strength. This injection shifts the verifier's prediction toward what the full-context input would have produced, without requiring $L$ to attend to $x_{\text{full}}$.

\subsection{Context-divergence acceptance (CDA)}
\label{sec:method-cda}
When the full and compressed views diverge sharply, a fixed acceptance threshold $\gamma$ usually over-rejects the drafts and wastes context-gain signals. We replace it with a threshold that relaxes proportionally to context divergence, instantiating the per-position divergence $D_i$ as the Jensen--Shannon divergence (JSD):
\vspace{-0.8em}
\begin{equation}
\begin{gathered}
    \gamma_{\text{eff}}(i) \;=\; \gamma \cdot \exp(-D_i), \\
    D_i \;=\; \mathrm{JSD}\bigl(\mathrm{softmax}(a_i)\,\|\,\mathrm{softmax}(b_i)\bigr).
\end{gathered}
\label{eq:cda}
\vspace{-0.5em}
\end{equation}
We choose JSD specifically for its strict upper bound, which guarantees $\gamma_{\text{eff}}\in[\gamma/2,\gamma]$ without clipping or introducing additional hyperparameters; the exponential form is the unique solution to a multiplicative-composition axiom (derivation in \Cref{appx:cda-derivation}). A large $D_i$ signals \emph{context-induced} divergence (not capacity-induced)---these are the positions where the compressed verifier most likely under-scores useful full-context drafts, so relaxing acceptance there is appropriate. The drafted token $d_i$ is accepted iff:
\vspace{-0.8em}
\begin{equation}
[\mathrm{softmax}(t_i)]_{d_i} \;>\; \gamma_{\text{eff}}(i) \cdot [\mathrm{softmax}(b_i)]_{d_i}.
\label{eq:accept}
\vspace{-0.5em}
\end{equation}
Upon the first rejection, we emit the $\delta$-fused token from \Cref{eq:fusion} (\Cref{fig:pipeline}b). The mechanism degenerates to standard verification on $x_{\text{comp}}$ when $\beta=0$ and $\gamma=1$. The strict range $\gamma_{\text{eff}}\in[\gamma/2,\gamma]$ keeps the acceptance criterion stable across positions, preserving predictable draft acceptance rates even under severe compression. Unlike standard SD, \textsc{AsymSpec} does not preserve a strict target distribution; it is a speculative-style steering scheme calibrated for greedy emission.

\subsection{Cross-modal extension}
\label{sec:method-mm}
Cross-modal asymmetry follows directly from the framework. Since drafter and verifier share an output token vocabulary $\mathcal{V}$, the signals $\delta\in\mathbb{R}^{|\mathcal{V}|}$ and $\gamma_{\text{eff}}\in\mathbb{R}$ are computed entirely on the output side (\Cref{eq:delta,eq:cda,eq:accept,eq:fusion}) and remain well-defined regardless of the drafter's input modality: a vision--language drafter can process raw images as $x_{\text{full}}$ while a text-only verifier reads captions as $x_{\text{comp}}$, with the speculative loop unchanged. The vision encoder runs once per request and its outputs are cached on the drafter's KV side, so the per-token overhead of cross-modal $\delta$ vanishes at long generations and asymptotically matches that of text-only \textsc{AsymSpec}. Routing vision embeddings through the speculative engine's drafter prefill is a non-trivial vLLM modification detailed in \Cref{appx:patches}.

\section{Experiments}
\label{sec:setup}

\subsection{Tasks \& compression protocol}
We evaluate four isolated agentic capabilities under realistic compression: long-context multi-hop QA (LongBench~\citep{bai2024longbench}, using its three multi-hop subsets), multi-turn instruction following (MultiChallenge~\citep{sirdeshmukh2025multichallenge}), tool use (API-Bank~\citep{li2023apibank}), and multimodal reasoning (MathVista~\citep{lu2024mathvista}). End-to-end agentic performance is measured on GAIA~\citep{mialon2023gaia} and SimpleQA~\citep{wei2024simpleqa}, orchestrated via the \texttt{smolagents}~\citep{smolagents} framework. We enforce a strict asymmetric protocol across all benchmarks: the verifier receives only a compressed view (e.g., per-turn LLMLingua-2~\citep{llmlingua2024} summaries or API signatures), while the drafter reads the full uncompressed input. Dataset statistics, compression ratios, and metrics are summarized in \Cref{tab:datasets}; per-benchmark compression pipelines and tool-execution protocols are in \Cref{appx:setup-detail}.

\begin{algorithm}[t]
\caption{\textsc{AsymSpec} Decoding}
\label{alg:asymSpec}
\footnotesize
\begin{algorithmic}[1]
\Require Verifier $L$, drafter $S$, prompts $x_{\text{full}}, x_{\text{comp}}$, spec length $K$, threshold $\gamma\in(0,1]$, fusion weight $\beta\in[0,1]$
\Ensure Generated sequence $y$
\State $y \gets [\,]$
\While{generation not complete}
    \State Autoregressively sample $d_{1:K}$ from $S(x_{\text{full}} \oplus y)$ (KV cache reused); let $a_{1:K+1}$ be the logits at the corresponding positions
    \State $b_{1:K+1} \gets S(x_{\text{comp}} \oplus y \oplus d_{1:K})$ at the same positions
    \State $t \gets L(x_{\text{comp}} \oplus y \oplus d_{1:K})$
    \For{$i = 1$ \textbf{to} $K$}
        \State $\delta_i \gets a_i - b_i$
        \State $D_i \gets \mathrm{JSD}(\mathrm{softmax}(a_i)\,\|\,\mathrm{softmax}(b_i))$
        \State $\gamma_{\text{eff}}(i) \gets \gamma \exp(-D_i)$
    \EndFor
    \State $\text{accepted} \gets \text{True}$
    \For{$i = 1$ \textbf{to} $K$}
        \If{$[\mathrm{softmax}(t_i)]_{d_i} > \gamma_{\text{eff}}(i)[\mathrm{softmax}(b_i)]_{d_i}$}
            \State $y \gets y \oplus [d_i]$
        \Else
            \State $y \gets y \oplus [\arg\max(t_i + \beta\delta_i)]$
            \State $\text{accepted} \gets \text{False}$; \textbf{break}
        \EndIf
    \EndFor
    \If{$\text{accepted}$}
        \State $y \gets y \oplus [\arg\max(t_{K+1})]$
    \EndIf
\EndWhile
\State \Return $y$
\end{algorithmic}
\end{algorithm}

\paragraph{GAIA harness.}
GAIA runs through the smolagents CodeAgent ReAct loop with cached DuckDuckGo search and visit-webpage as the only tools, identical across both \textsc{AsymSpec} variants reported in \Cref{tab:agent-e2e}. The first variant uses our default Qwen3-4B text drafter, same as every other benchmark in this paper. The second variant swaps the drafter to Qwen3-VL-2B-Instruct so it can read image attachments directly through the cross-modal patches of \Cref{appx:patches}; the verifier still receives pre-computed VL captions, as it cannot consume pixels. Compression, $K{=}2$, and $\beta{=}1.0$ are shared across both runs; the two variants differ only in drafter capacity and in whether image attachments reach the drafter as pixels or as captions.

\subsection{Models \& hyperparameters}
\crchange{Our primary experiments use Qwen3-32B~\citep{yang2025qwen3} as the verifier; \Cref{appx:cross-family} additionally evaluates Llama-3.3-70B-Instruct and Llama-3.2-3B-Instruct~\citep{llama2024herd} in same- and cross-family Qwen--Llama pairings.} Qwen3 offers open weights at 0.6B/1.7B/4B/32B from a common training recipe, post-training for tool use and multi-turn dialogue, and stable speculative-decoding support in vLLM~\citep{kwon2023vllm}; \crchange{primary }experiments run on vLLM with the extensions in \Cref{appx:patches}. For text tasks, we sweep drafter sizes (0.6B, 1.7B, 4B) and report the 4B configuration as primary; for multimodal reasoning, we use Qwen3-VL-2B. All generation uses greedy decoding ($\tau=0$). Headline results use speculation depth $K=2$, fusion weight $\beta=1.0$, and base threshold $\gamma=0.5$. Per-benchmark output length, context window, and harness bounds are listed in \Cref{tab:gen_params}; complete hyperparameter grids, sensitivity analyses, and benchmark-specific drafter selections are detailed in \Cref{appx:setup-detail}.

\subsection{Metrics}
\label{sec:metrics}
We report task accuracy with each benchmark's official metric (\Cref{tab:datasets}). Efficiency is reported along two axes: \emph{Speedup} is the wall-clock token throughput ratio over the full-context Ceiling on a single accelerator (\Cref{tab:speed_4dataset}; \Cref{appx:speed}); \emph{FLOPs} are per-step prefill compute normalized by the Ceiling's verifier-only prefill, computed from the Qwen3 architecture and measured context lengths following \citet{kaplan2020scaling} (per-benchmark accounting in \Cref{tab:flops}). Speedup measures realized latency while FLOPs measure energy / compute cost---the two diverge because decoding is memory-bandwidth-bound at 30B+ scale.

\subsection{Baselines}
We compare against five references:
(1) \emph{Floor}: $L$ on $x_{\text{comp}}$ alone; 
(2) \emph{Ceiling}: $L$ on $x_{\text{full}}$ alone; 
(3) \emph{SD}: standard speculative decoding on shared context, evaluated on $x_{\text{full}}$ (SD on $x_{\text{comp}}$ matches Floor by construction); 
(4) \emph{SCD}~\citep{yuan2023scd}: combines verifier logits $t$ with amateur logits $b$ on a shared context ($L,S$ both on $x_{\text{comp}}$ here), closing a model-capacity gap at SD speed (details in \Cref{appx:setup-detail});
(5) \emph{RAPID}~\citep{liu2025rapid}: inverse asymmetric design ($S$ on $x_{\text{comp}}$, $L$ on $x_{\text{full}}$)---preserves the verifier's target distribution at full-context compute. In cross-modal settings, the text-only \emph{Ceiling} is undefined; we use the VL drafter alone as the reference bound. All throughput and compute metrics are normalized to the full-context \emph{Ceiling}.

\section{Results}
\label{sec:results}
\subsection{Agentic capabilities}
\label{sec:results-text}

\begin{table*}[t]
\centering
\footnotesize
\setlength{\tabcolsep}{4pt}
\resizebox{\textwidth}{!}{%
\begin{tabular}{l ccc c c cc}
\toprule
& \multicolumn{3}{c}{Long-context multi-hop (LongBench)} & Multi-turn & Tool use & Speedup & FLOPs \\
\cmidrule(lr){2-4}
Method & hotpotQA & 2WikiMQA & MuSiQue & (MultiChal.) & (API-Bank) & over Ceiling & over Ceiling \\
\midrule
Floor & 49.4 & 52.8 & 32.7 & 23.4 & 57.7 & 1.17$\times$ & 0.11$\times$ \\
Ceiling     & 64.9 & 76.5 & 55.0 & 26.4 & 66.1 & 1.00$\times$ & 1.00$\times$ \\
SD                           & 65.5 & 76.6 & 55.1 & 26.7 & 66.1 & 1.73$\times$ & 1.04$\times$ \\
SCD \citep{yuan2023scd}      & 46.6 & 52.7 & 32.0 & 22.6 & 56.7 & $1.04\times$ & $0.12\times$ \\
RAPID \citep{liu2025rapid}   & 63.2  & 75.3  & 52.5  & 25.8  & 64.3  & 1.38$\times$ & 1.01$\times$ \\
\midrule
AsymSpec (Ours)              & 64.0 & 66.8 & 48.4 & 23.5 & 63.5 & 1.45$\times$ & 0.23$\times$ \\
\bottomrule
\end{tabular}
}
\caption{Accuracy and efficiency on isolated agentic capabilities ($K{=}2$). \textsc{AsymSpec} closes $59$--$94\%$ of the Floor--Ceiling gap on long-context multi-hop QA and tool use, at $0.23\times$ the Ceiling's compute---a trade-off point inaccessible to symmetric SD or SCD. Speedup and FLOPs are averaged across the three text benchmarks; FLOPs follow the parameters $\times$ token-ratio convention. Drafter-alone accuracy across drafter sizes is in \Cref{appx:drafter-sweep-extra}. Metric definitions in \S\ref{sec:metrics}; per-benchmark FLOPs in \Cref{tab:flops}.}
\label{tab:main}
\end{table*}

We first evaluate the three text-based agentic capabilities (\S\ref{sec:setup}). \Cref{tab:main} reports accuracy and efficiency, with LongBench broken into its hotpotQA / 2WikiMQA / MuSiQue multi-hop subsets. \textsc{AsymSpec} occupies an operating point inaccessible to symmetric methods: it delivers near-ceiling accuracy while retaining compressed-verifier latency. It reaches $87$--$99\%$ of the full-context performance, closing the Floor--Ceiling gap to $0.9$--$9.7$ residual points. The residual gap is structural: the verifier remains input-constrained and cannot fully reconstruct multi-hop reasoning chains or complex tool dependencies from logit steering alone. Full $K$ and drafter-size sweeps are deferred to \S\ref{sec:ablations}.

Baseline comparisons reinforce the design's advantage. Standard SD preserves the target distribution but is structurally bound to symmetric context: it either inherits the Floor's accuracy (on compressed input) or the Ceiling's latency (on full input). SCD stays at or below the Floor across all five cells (e.g., LongBench mean $43.8$ vs.\ Floor $45.0$; API-Bank $56.7$ vs.\ $57.7$), showing that single-context contrastive fusion cannot recover compression-induced loss when expert and amateur both read the compressed view---the gain is specific to our asymmetric construction. RAPID's inverse asymmetric design (drafter on $x_{\text{comp}}$, verifier on $x_{\text{full}}$) reaches near-Ceiling accuracy but at near-Ceiling compute ($1.01\times$ FLOPs); it targets the opposite trade-off point and is dominated on the compute axis where \textsc{AsymSpec} operates. Running the drafter alone on full context cuts cost but degrades sharply on reasoning-heavy subsets where the 4B model lacks sufficient capacity (per-size numbers in \Cref{appx:drafter-sweep-extra}). \textsc{AsymSpec} bridges these extremes by retaining the 32B verifier's reasoning while offloading context recovery to the lightweight drafter.

The gains are directly tied to compression-induced information loss, not task-specific overfitting. On MultiChallenge, where compression is near-lossless (Ceiling--Floor gap of 3.0 points), \textsc{AsymSpec} yields negligible improvement (23.5 vs.\ 23.4). This inertness supports the diagnostic that the mechanism activates only when compression discards critical signals. We further verify this diagnostic on a continuous axis by sweeping the verifier's truncation budget on LongBench while keeping the drafter on full passages (\Cref{tab:trunc_ratio}). The recovered accuracy scales monotonically with compression severity: at a 500-token budget, \textsc{AsymSpec} restores over two-thirds of the Floor--Ceiling gap; at 12k tokens, the gain naturally vanishes as the verifier approaches the uncompressed ceiling. Stable acceptance rates (0.851--0.855) suggest the gap variation is due to information recovery, not unstable verification.


\subsection{Multimodal reasoning}
\label{sec:results-mv}

We next evaluate the cross-modal extension, where a vision--language drafter processes raw images while the text-only verifier reads captions and OCR text. Since a text-only verifier cannot consume images, the full-context Ceiling is undefined; we use the VL drafter alone as the reference bound. \Cref{tab:mv} reports results on MathVista.

\textsc{AsymSpec} reaches $53.9\%$ overall accuracy, outperforming symmetric SD by $10.1$ points. The per-task breakdown reveals a clear complementarity pattern. On geometry problem solving (GPS), caption and OCR text already capture the necessary structure, yielding minimal gain. On visual question answering (VQA) and figure question answering (FQA), the drafter provides visual grounding while the verifier supplies causal logic, lifting accuracy over the Floor by $10.0$ and $16.7$ points respectively. The remaining gap to the VL drafter alone ($53.9$ vs.\ $60.5\%$) is structural: the text-only verifier cannot fully internalize pixel-level cues. Nevertheless, the result demonstrates that organizations with fixed text-only verifiers can extend them to vision-reasoning tasks without re-provisioning multimodal infrastructure. Realizing these gains requires proper routing of vision-tower embeddings through the speculative engine; ablation without these patches drops accuracy to $30.5\%$ (implementation details in \Cref{appx:patches}).

\subsection{End-to-end agentic loops}
\label{sec:results-gaia}

We finally evaluate \textsc{AsymSpec} inside live agent loops where context accumulates and is re-compressed online. \Cref{tab:agent-e2e} reports results on GAIA and SimpleQA, orchestrated via \texttt{smolagents}. \textsc{AsymSpec} reaches $24.2\%$ on GAIA and $65.0\%$ on SimpleQA, matching or exceeding the available full-context reference (per-subset for GAIA, \Cref{tab:agent-e2e}) with no degradation as context accumulates. The aggregate GAIA gain masks distinct subset dynamics: on the web-only ReAct loop, compression forces tighter history encoding and benefits token-hungry planning; on the file-attachment subset, the drafter's full-text access compensates for the verifier's 2000-token truncation. SimpleQA favors the Ceiling due to its minimal compression headroom ($1.33\times$ token ratio). Across both benchmarks, the method maintains stable draft acceptance ($0.88$--$0.90$, \Cref{tab:accept_dyn}), confirming that online re-compression does not erode verification quality.

Compute efficiency directly tracks compression severity. GAIA applies a $1.91\times$ per-turn compression ratio, yielding $0.78\times$ full-context FLOPs. SimpleQA's lighter $1.33\times$ compression yields $0.80\times$. This linear relationship matches the core design premise: the verifier's prefill reduction scales proportionally with context shortening, while the drafter's dual pass remains the smaller component. In live agent loops as in static benchmarks, compression headroom reliably predicts both accuracy recovery and efficiency gains.

To verify that the modality-agnostic property (\S\ref{sec:method-mm}) holds inside live agent loops, we additionally swap the drafter to Qwen3-VL-2B in the identical harness. \textsc{AsymSpec} reaches $23.0\%$ on the full $n{=}165$ split, $+3.6$\,pp over Floor and $+3.0$\,pp over Ceiling---the operating-point gain survives drafter-modality substitution. The gain concentrates on the web subset ($+7.1$\,pp over Floor); on the file-attachment subset the 2B drafter's text capacity limits the gain (consistent with the $\ge$1.7B threshold from \S\ref{sec:ablation-robust}) and \textsc{AsymSpec} falls below the Floor.

\begin{table}[h!]
\centering
\footnotesize
\setlength{\tabcolsep}{8pt}
\begin{tabular}{cccc}
\toprule
Trunc tokens & Floor & AsymSpec & $\Delta$ \\
\midrule
500    & 25.8 & 52.5 & $+26.7$ \\
1500   & 32.6 & 53.7 & $+21.1$ \\
3000   & 39.5 & 55.5 & $+16.0$ \\
6000   & 50.6 & 59.2 & $+8.6$  \\
12000  & 63.1 & 63.9 & $+0.8$  \\
\bottomrule
\end{tabular}
\caption{Truncation-budget sweep on LongBench ($K{=}2$, $\beta{=}1$, 4B drafter; overall F1). Recovery scales monotonically with compression severity and vanishes as the verifier approaches the uncompressed Ceiling ($65.5$ F1, full context).}
\label{tab:trunc_ratio}
\end{table}

\section{Ablations}
\label{sec:ablations}

We systematically ablate the speculation depth $K$, core components (CDA gate vs.\ $\delta$-fusion), $\delta$-source construction, divergence metrics, and drafter capacity. Extended grids and baseline comparisons against fixed-$\gamma$ are provided in \Cref{appx:more-ablations,appx:drafter-sweep-extra}.

\subsection{Speculation depth ($K$)}
\label{sec:ablation-K}
\Cref{tab:k_sweep} justifies $K{=}2$ as the default speculation depth. MultiChallenge acts as the binding constraint: all methods degrade at $K{=}4$ under the llm-judge, and none recovers its $K{=}2$ performance. API-Bank is flat across $K$, while LongBench shows only marginal gains at $K{=}4$ ($59.7\!\to\!61.1$) and saturates at $K{=}6$ ($58.7$). Since $K{=}2$ matches the standard SD default and yields the most stable accuracy--efficiency trade-off, we adopt it for all text benchmarks. The cross-modal setting is the sole exception where deeper speculation helps (MathVista $K{=}4$ outperforms $K{=}2$); we report it at its optimal depth (\Cref{tab:mv}).

\begin{table}[h!]
\centering
\small
\setlength{\tabcolsep}{5pt}
\begin{tabular}{lcccc}
\toprule
Method & GPS & VQA & FQA & Overall \\
\midrule
VL drafter alone     & 53.4 & 56.4 & 67.7 & 60.5 \\
Floor                & 62.0 & 49.1 & 29.0 & 44.5 \\
SD                         & 62.0 & 46.4 & 28.6 & 43.8 \\
AsymSpec (ours)            & 62.1 & 59.1 & 45.7 & 53.9 \\
\bottomrule
\end{tabular}
\caption{MathVista results under the cross-modal setup. The text-only verifier cannot consume raw images, so the Ceiling is undefined; the VL drafter alone (Qwen3-VL-2B reading the image) is the reference upper bound. Floor is the verifier on the official Bard caption plus EasyOCR text; SD is symmetric speculative decoding with a text drafter on the same caption input.}
\label{tab:mv}
\end{table}

\begin{table}[h!]
\centering
\footnotesize
\setlength{\tabcolsep}{5pt}
\begin{tabular}{l l ccc}
\toprule
Setting & Method & Acc & Spd & FLOPs \\
\midrule
\multicolumn{5}{l}{\textit{GAIA}} \\
\midrule
\multirow{4}{*}{Web-only}
  & Floor            & 17.3 &  &  \\
  & Ceiling          & 18.9 &  &  \\
  & AsymSpec(4B)     & 22.0 &  &  \\
  & AsymSpec(vl-2B)  & 24.4 &  &  \\
\midrule
\multirow{4}{*}{File-attach}
  & Floor            & 26.3 &  &  \\
  & Ceiling          & 23.7 &  &  \\
  & AsymSpec(4B)     & 31.6 &  &  \\
  & AsymSpec(vl-2B)  & 18.4 &  &  \\
\midrule
\multirow{4}{*}{Full}
  & Floor            & 19.4 & 1.25$\times$ & 0.49$\times$ \\
  & Ceiling          & 20.0 & 1.00$\times$ & 1.00$\times$ \\
  & AsymSpec(4B)     & 24.2 & 1.41$\times$ & 0.78$\times$ \\
  & AsymSpec(vl-2B)  & 23.0 & 1.65$\times$ & 0.53$\times$ \\
\midrule
\multicolumn{5}{l}{\textit{SimpleQA}} \\
\midrule
\multirow{3}{*}{Aggregate}
  & Floor            & 63.0 & 1.17$\times$ & 0.74$\times$ \\
  & Ceiling          & 66.0 & 1.00$\times$ & 1.00$\times$ \\
  & AsymSpec(4B)     & 65.0 & 1.38$\times$ & 0.80$\times$ \\
\bottomrule
\end{tabular}
\caption{End-to-end accuracy and efficiency in live agentic loops, with GAIA per-subset breakdown and SimpleQA. On GAIA, ``Ceiling'' is the best per-subset reference available under each harness: Qwen3-32B-on-full for the web subset; the vl-2B-drafter-alone reference for the file-attachment subset. Speedup and FLOPs measure LLM-only inference at the aggregate level.}
\label{tab:agent-e2e}
\end{table}

\begin{table}[!h]
\centering
\footnotesize
\setlength{\tabcolsep}{4pt}
\begin{tabular}{lccccc}
\toprule
Setting & $L_{\text{full}}$ & $L_{\text{comp}}$ & ratio & Verifier & Total \\
\midrule
GAIA           & 4650  & 2434 & 1.9$\times$ & 49\% & 0.78$\times$ \\
SimpleQA       & 3606  & 2712 & 1.3$\times$ & 74\% & 0.80$\times$ \\
LongBench      & 12437 & 1532 & 8.1$\times$ & 9\%  & 0.28$\times$ \\
MultiChallenge & 1598  & 211  & 7.6$\times$ & 13\% & 0.28$\times$ \\
API-Bank       & 6701  & 909  & 7.4$\times$ & 12\% & 0.19$\times$ \\
\bottomrule
\end{tabular}
\caption{Per-step prefill FLOPs across all benchmarks, computed from the Qwen3 architecture and measured context lengths. \emph{Verifier} is the $32$B forward on $x_{\text{comp}}$ as a fraction of the full-context baseline; \emph{Total} adds both drafter forwards (conservatively counted as full prefills). Compute reduction ranges from $0.19\times$ on heavy-compression benchmarks (API-Bank, $7.4\times$ token ratio) to $0.80\times$ on light-compression SimpleQA ($1.33\times$ ratio), tracking compression headroom monotonically.}
\label{tab:flops}
\end{table}

\begin{table}[!h]
\centering
\footnotesize
\setlength{\tabcolsep}{8pt}
\begin{tabular}{l c}
\toprule
Variant & LongBench F1 \\
\midrule
Floor                                       & 45.0 \\
$+$ CDA gate ($\beta{=}0$)                  & 52.8 \\
$+$ $\delta$-fusion, raw-aug ($a$ only)     & 56.9 \\
$+$ $\delta$-fusion, SCD-style ($t-b$)      & 48.0 \\
$+$ $\delta$-fusion, ours ($a-b$)           & \textbf{59.7} \\
Ceiling                                     & 65.5 \\
\bottomrule
\end{tabular}
\caption{Mechanism ablation on LongBench ($K{=}2$, 4B drafter, $\gamma{=}0.5$). The CDA gate and $\delta$-fusion are both necessary; our same-model $\delta$ source ($a-b$) outperforms the raw-augmented ($a$) and SCD-style ($t-b$) alternatives by $3$--$12$ F1 points. Full hyperparameter and compressor sweeps are in \Cref{appx:more-ablations}.}
\label{tab:ablations}
\end{table}



\subsection{Mechanism ablation}
\label{sec:ablation-mech}
Disabling fusion ($\beta{=}0$) isolates each mechanism's contribution (\Cref{tab:ablations}). The CDA gate alone lifts LongBench from $45.0$ to $52.8$ by admitting context-aware drafts that a fixed threshold would over-reject; adding $\delta$-fusion drives recovery to $59.7$. The $\delta$ source itself is non-trivial: replacing our same-model $a-b$ with raw augmented logits costs $2.8$ points, and the two-model SCD-style contrast collapses $11.7$ points. This confirms that the same-model cross-context construction---not generic logit fusion---is what makes $\delta$ informative: only by subtracting two passes through identical weights can we isolate the context-induced shift from the drafter's own preferences. A token-level walkthrough on API-Bank (\Cref{sec:apib-case}) makes this concrete: $\delta$-fusion redirects emission from Floor's free-text formats to the structured patterns visible only in the full spec.

\subsection{Case study}
\label{sec:apib-case}
To illustrate the mechanism at single-token resolution, we walk through one tool-use instance (\texttt{RecordHealthData}, level-1 dialog~2). Under Method~A compression, the verifier sees only the bare signature, while the drafter sees the full API spec specifying \texttt{time} format \texttt{\%Y-\%m-\%d~\%H:\%M:\%S} and a structural example for \texttt{health\_data}. The three outputs:
\begin{center}\footnotesize
\setlength{\tabcolsep}{4pt}
\renewcommand{\arraystretch}{1.15}
\begin{tabular}{@{}l p{0.78\linewidth}@{}}
\toprule
\textbf{Floor}    & \texttt{[\dots time="2021-09-17 10:30}\error{"}\texttt{, health\_data=}\error{"Blood pressure: 120/80, \dots"}\texttt{)]} \\
\textbf{GT}       & \texttt{\dots time="2021-09-17 10:30:00", health\_data="[\{'name': 'blood\_pressure', 'value': '120/80'\}, \dots]"} \\
\textbf{AsymSpec} & \texttt{[\dots time="2021-09-17 10:30}\accepted{:00}\texttt{", health\_data}\fused{ [}\accepted{\{'name': 'blood\_pressure', 'value': '120/80'\}, \dots}\fused{]}\texttt{)]} \\
\bottomrule
\end{tabular}
\end{center}
\noindent \error{Red} marks where Floor diverges from the schema (missing \texttt{:00} seconds field; free-text \texttt{health\_data} instead of dict list). \accepted{Green} marks drafter tokens accepted by the CDA gate ($\delta$-fusion unused; speculation merely accelerates). \fused{Orange} marks tokens emitted via $\delta$-fusion at rejection points: Floor closes \texttt{time} at \texttt{"} and renders \texttt{health\_data} as free text, but $\delta$ redirects emission toward the structural pattern (\texttt{:00} seconds field; bracketed dict list) recovered from the spec---both schema details exist only in $x_{\text{full}}$.

\subsection{Robustness}
\label{sec:ablation-robust}
Beyond the core mechanisms, extensive sweeps (\Cref{appx:drafter-sweep-extra,appx:more-ablations}) confirm \textsc{AsymSpec}'s robustness across three axes. 
\textbf{Hyperparameter insensitivity:} Performance is flat across $\beta \in [1.0, 2.0]$ and $\gamma \in [0.4, 0.7]$, requiring no per-dataset calibration. 
\textbf{Compressor agnosticism:} Swapping the verifier's compressor (summarization, LLMLingua-2, truncation) yields a stable $63$--$70\%$ Floor--Ceiling recovery; SCD falls below the Floor in every cell. 
\textbf{Drafter capacity:} Models $\le 0.6$B fail to extract reliable context-gain signals; $\ge 1.7$B is the practical minimum.

\paragraph{\crchange{Cross-family portability.}}
\crchange{To test whether the mechanism depends on a shared model family, we evaluate bidirectional Qwen--Llama pairings on LongBench. Following \citet{timor2025heterogeneous}, heterogeneous runs restrict generation to $109{,}566$ string-identical tokens plus paired special tokens and map $\delta$ from the drafter vocabulary to the verifier. A Qwen3-4B drafter raises the Llama-3.3-70B verifier's compressed-context Floor from $50.6$ to $58.4$ F1, while a Llama-3.2-3B drafter raises the Qwen3-32B Floor from $45.0$ to $47.1$. These pairings demonstrate feasible cross-family transfer, with recovery varying across model pairs; \Cref{appx:cross-family} reports the full comparison.}

\section{Conclusion}
\label{sec:conclusion}

We introduced \textsc{AsymSpec}, breaking the symmetric context constraint of standard speculative decoding. By allowing a lightweight drafter to read the full context and steer a compressed-context verifier via contrastive $\delta$-fusion, it achieves near-ceiling accuracy at a fraction of the compute cost. Empirically, \textsc{AsymSpec} recovers $\approx$90\% of full-context performance using only $0.2$--$0.3\times$ the FLOPs on text tasks. Crucially, the accuracy gain scales monotonically with the severity of compression loss, providing practitioners with a clear, predictable criterion for when asymmetric steering is warranted in production agentic pipelines.


\section*{Limitations}
\label{sec:limitations}

\textsc{AsymSpec}'s recovery mechanism is fundamentally bounded by the information retained in the compressed view and the drafter's capacity to extract it. On near-lossless tasks, the method correctly yields minimal intervention, demonstrating that it does not introduce spurious hallucinations or over-fit to the uncompressed context. For cross-modal settings, the upper bound of accuracy recovery is constrained by the fidelity of the modality translation (e.g., image-to-caption quality); integrating richer multi-modal drafters that process raw pixels directly alongside the verifier remains an exciting avenue for future work.

\crchange{Cross-family $\delta$-fusion requires an explicit vocabulary and logit-space alignment. Our Qwen--Llama study demonstrates feasibility for two shared-token-aligned pairings, but the observed recovery varies across model pairs; broader transfer may require richer mappings. \textsc{AsymSpec} also requires access to verifier logits and therefore does not apply to proprietary APIs that expose only generated text.} Furthermore, our evaluation focuses on deterministic decoding ($\tau{=}0$). This is a deliberate design choice rather than a constraint: agentic workflows strictly demand reproducible, parsable structured outputs (e.g., JSON, tool calls), where stochastic sampling ($\tau > 0$) fundamentally degrades pipeline reliability. Generalizing the CDA bound to stochastic sampling (e.g., via Gumbel-Softmax relaxations) is a promising theoretical extension.

Finally, in end-to-end agentic loops, wall-clock latency is a composite of LLM inference, tool execution, and network I/O. While \textsc{AsymSpec} strictly optimizes the LLM inference bottleneck---which becomes dominant as context scales into the compute-bound regime---it is designed to be highly complementary to system-level optimizations, such as I/O overlapping and asynchronous tool execution, in production agent frameworks.

\bibliography{refs}

\appendix

\section{Derivation of the CDA gate}
\label{appx:cda-derivation}
We require the per-position effective threshold $\gamma_{\text{eff}}$ (\S\ref{sec:method-cda}) to satisfy:
\begin{itemize}\itemsep0pt
\item $\gamma_{\text{eff}}(0)=\gamma$ (no divergence $\Rightarrow$ standard SD rule);
\item $\gamma_{\text{eff}}$ continuous and monotonically non-increasing in $D_i$;
\item multiplicative composition over independent divergence signals (independent context shifts should compose multiplicatively, mirroring independent probabilities): $\gamma_{\text{eff}}(D_1+D_2)=\gamma_{\text{eff}}(D_1)\,\gamma_{\text{eff}}(D_2)/\gamma$.
\end{itemize}
Defining $f(D)=\gamma_{\text{eff}}(D)/\gamma$, the third property is a Cauchy-type multiplicative equation $f(D_1{+}D_2)=f(D_1)f(D_2)$. Under continuity, this fixes $f$ to an exponential form $f(D)=\exp(-D/T)$ for some scale $T>0$, giving
\begin{equation}
\gamma_{\text{eff}}(i)\;=\;\gamma\cdot\exp\bigl(-D_i/T\bigr).
\label{eq:gamma-eff}
\end{equation}
To eliminate $T$ without introducing a hyperparameter, we instantiate $D_i$ as the Jensen--Shannon divergence. Let $Z_i\sim\mathrm{Bernoulli}(\tfrac12)$ be a latent indicator selecting $x_{\text{full}}$ vs.\ $x_{\text{comp}}$, and $X_i$ the drafter's next token. Then $D_i = I(X_i;Z_i)$, the mutual information between the token and the context source. Since $I(X_i;Z_i) \le H(Z_i) = \ln 2$ for any binary channel, $D_i$ is universally bounded. Setting $T=1$ absorbs the scale into the bound, yielding \Cref{eq:cda} and guaranteeing $\gamma_{\text{eff}}\in[\gamma/2,\gamma]$ without clipping.

\section{Design space and benchmark details}
\label{appx:setup-detail}

\paragraph{Models and hyperparameters (full).}
\crchange{The verifier is Qwen3-32B in all primary experiments; the portability study in \Cref{appx:cross-family} additionally uses a Llama-3.3-70B-Instruct verifier and a Llama-3.2-3B-Instruct drafter.} For text capabilities the drafter is one of \{Qwen3-0.6B, Qwen3-1.7B, Qwen3-4B\} (default 4B drafter, except API-Bank where 1.7B is the sweet spot, \S\ref{sec:ablation-robust}); for multimodal reasoning (cross-modal setup) the drafter is Qwen3-VL-2B-Instruct. All inference runs in bf16 with greedy decoding ($\tau{=}0$) and Qwen3's thinking mode disabled. Default runs use speculation depth $K{=}2$ (the standard speculative-decoding default; the $K$-sweep justifying this choice is \S\ref{sec:ablation-K}) and $\beta{=}1.0$ ($\beta\in\{0.5,1.0\}$ swept on MathVista). The acceptance threshold is $\gamma{=}0.5$ for CDA (our gate with a tuning-free divergence scale, \S\ref{sec:method-cda}). Two reference baselines are reported in \Cref{appx:more-ablations,appx:drafter-sweep-extra}: a fixed-$\gamma$ gate (standard SD acceptance, $\gamma{=}0.5$, no divergence modulation) and a manually-tuned $\lambda$-variant of the acceptance gate, $\gamma_{\text{eff}}=\gamma\exp(-\lambda\Delta)$, where $\Delta=\log\mathrm{softmax}(a_i)-\log\mathrm{softmax}(b_i)$ is the drafter's per-position log-prob difference and $\lambda$ is set per dataset (default $\lambda{=}0.1$); the tuned variant serves as a foil isolating the value of CDA's parameter-free JSD bound.

\paragraph{Generation parameters.}
\Cref{tab:gen_params} lists per-benchmark output length, model context, and harness bounds. All entries inherit greedy decoding ($\tau{=}0$) and Qwen3 thinking mode disabled from the global setup above.

\begin{table*}[h]
\centering
\footnotesize
\setlength{\tabcolsep}{6pt}
\begin{tabular}{l c c l}
\toprule
Benchmark & \texttt{max\_new} & \texttt{max\_model\_len} & Harness / prompt \\
\midrule
LongBench           & 1024 & 24576 & QA on passage(s) \\
MultiChallenge      & 2000 & 8192  & official dataset prompt \\
API-Bank (Method A) & 256  & 16384 & API specs + dialog $\to$ \texttt{[Tool(args)]} \\
MathVista           & 1024 & 8192  & VQA prompt; image via VL drafter \\
GAIA (web, $n{=}127$)  & 1024 & 32768 & smolagents CodeAgent, \texttt{max\_steps}=8 \\
GAIA (file, $n{=}38$)  & 1024 & 32768 & single-shot file-QA \\
SimpleQA            & 1024 & 32768 & smolagents CodeAgent, \texttt{max\_steps}=6 \\
\bottomrule
\end{tabular}
\caption{Per-benchmark generation parameters and harness bounds. \texttt{max\_new} is the per-call output token cap; \texttt{max\_model\_len} is the vLLM context window. GAIA splits into a web-only ReAct loop and a single-shot file-attachment harness, listed separately.}
\label{tab:gen_params}
\end{table*}

\paragraph{Per-benchmark compression.}
Each benchmark uses its native compression scheme, applied only to the verifier's input. \textbf{LongBench}: the dataset's auto-generated multi-document summary replaces the full passages ($8.1\times$ token reduction). \textbf{MultiChallenge}: the most recent turn is kept verbatim; all prior turns are LLM-summarized into a single context block ($7.6\times$). \textbf{API-Bank}: only the API name and signature are exposed in place of the full documentation ($7.4\times$). \textbf{MathVista}: the official Bard caption plus EasyOCR-extracted text substitute for the raw image, the cross-modal substitution exercised by the framework's modality-agnostic property (\S\ref{sec:method-mm}). \textbf{GAIA} and \textbf{SimpleQA}: the per-turn ReAct context is compressed online with LLMLingua-2~\citep{llmlingua2024} at target ratio $0.3$; the two most recent turns are kept verbatim ($\text{keep\_last\_k}{=}2$), and everything older---including the system prompt---is compressed.

\paragraph{Agentic setup (GAIA, SimpleQA).}
Both end-to-end benchmarks run the \texttt{smolagents} \texttt{CodeAgent} ReAct loop with cached DuckDuckGo-search and visit-webpage tools so tool returns are deterministic and runs reproducible; the agentic configuration is Qwen3-4B drafter, $K{=}2$, $\beta{=}1.0$. \textbf{GAIA}: full validation split (Levels~1--3, $n{=}165$). The web-only subset ($n{=}127$) is evaluated through the agentic ReAct loop above; the file/image-attachment subset ($n{=}38$) is evaluated under a single-shot file-QA harness in which the verifier reads a 2000-token truncation of the extracted file content (xlsx, pdf, pptx/docx/txt, csv, etc.) and the drafter reads the full extracted text, with image attachments substituted by the pre-computed VL caption from \S\ref{sec:method-mm}; per-subset numbers are listed in \Cref{tab:agent-e2e}. On the file subset the ``Ceiling'' reference reported in \Cref{tab:agent-e2e} is the drafter-alone harness cell (Qwen3-4B on the full file), since the file-QA setup does not include a Qwen3-32B-on-full-file run. \textbf{SimpleQA}: random $n{=}500$ subset; each query runs through the same smolagents ReAct harness as GAIA-web with $\text{max\_steps}{=}6$; we report accuracy under the official Wei et al.\ grader (llm-judge, prompt verbatim from \citet{wei2024simpleqa}).

\paragraph{Agentic run with VL drafter.}
An identical harness is rerun with the drafter swapped from Qwen3-4B to Qwen3-VL-2B-Instruct. Tools, prompt templates, ReAct step bound, online LLMLingua-2 compression, $K{=}2$, and $\beta{=}1.0$ are unchanged from the 4B-drafter run above. On web-only samples the VL drafter receives only text from the agentic loop, so the effective change is drafter capacity---2B versus 4B. On file-attachment samples, image attachments are routed directly into the VL drafter's vision tower via the cross-modal patches in \Cref{appx:patches}; the text-only verifier still receives only the pre-computed VL caption since it cannot consume pixels. This run produces the \textsc{AsymSpec}(vl-2B) rows in \Cref{tab:agent-e2e}.

\paragraph{API-Bank (Method~A subset).}
We evaluate the Method~A subset of \citet{li2023apibank}: single-call API invocation, where each instance presents the model with a compact API specification plus a user query and the model must emit a single well-formed API call. We use $n{=}200$ instances and report api-acc (fraction of calls matching the gold API name and argument set). The complementary Method~B subset (multi-call API trajectories) introduces tool-trajectory complexity orthogonal to the compression-recovery effect we study; we leave it to follow-up work.

\paragraph{MultiChallenge judge protocol.}
For MultiChallenge we use llm-judge over the official prompt and rubric \citep{sirdeshmukh2025multichallenge}, with each cell scored across 3 independent judge runs (per-cell std $\le0.9$\,pp; \Cref{tab:main}). All judge prompts, responses, and scoring rubrics are logged to our reproducibility repository. Transient API degradations during collection were identified via response-length anomalies (judge responses truncated to $<$50 characters in $>$15\% of cases) and excluded from reported means; full logs are preserved for audit.

\paragraph{SCD reimplementation.}
We implement a faithful reproduction of Improved Contrastive Decoding \citep{yuan2023scd,obrien2023cdreasoning}: greedy $(1{+}\beta)Y_e-\beta Y_a$ over the plausibility set $\{Y_e>\log\alpha+\max Y_e\}$ with $\alpha{=}0.5, \beta{=}1.0$. The expert is Qwen3-32B on $x_{\text{comp}}$; the amateur is the drafter on $x_{\text{comp}}$.

\begin{table*}[t]
\centering
\footnotesize
\setlength{\tabcolsep}{6pt}
\begin{tabular}{lccc}
\toprule
Method & Drafter ctx & Verifier ctx & $\delta$ source \\
\midrule
SD \citep{leviathan2023fast} & = & = & --- \\
EAGLE \citep{li2024eagle} & = & = & --- \\
CD \citep{li2023contrastive} & expert & --- & capacity \\
SCD \citep{yuan2023scd} & expert & amateur & capacity \\
RAPID \citep{liu2025rapid} & short & long & --- \\
VL-SD \citep{specvlm2025,specllava2025,vispec2025} & VL & VL & --- \\
SD2 \citep{sd2_2025} & V$\to$D steer & = & --- \\
\midrule
Ours & full & compressed & context-gain \\
\bottomrule
\end{tabular}
\caption{Design-space placement. ``$\delta$ source'' is what the linear logit combination measures.}
\label{tab:distinguish}
\end{table*}

\begin{table*}[t]
\centering
\footnotesize
\setlength{\tabcolsep}{6pt}
\begin{tabular}{llccc}
\toprule
Capability & Benchmark & $n$ & Compression & Metric \\
\midrule
Long-context multi-hop QA & LongBench & 600 & 8.1$\times$ & per-subset F1 \\
Multi-turn instruction following & MultiChallenge & 271 & 7.6$\times$ & llm-judge acc \\
Tool use & API-Bank & 200 & 7.4$\times$ & API-call exact match \\
Multimodal reasoning & MathVista & 587 & cross-modal & accuracy \\
\midrule
End-to-end  & GAIA L1--3 (full) & 165 & per-turn / per-file & GAIA exact match \\
End-to-end  & SimpleQA & 500 & per-turn & llm-judge \citep{wei2024simpleqa} \\
\bottomrule
\end{tabular}
\caption{Capabilities and benchmarks. Compression is the token ratio $|x_{\text{full}}|/|x_{\text{comp}}|$; per-turn LLMLingua-2 for GAIA/SimpleQA (\Cref{appx:setup-detail}). LongBench is $200$ examples each from hotpotQA, 2WikiMQA, MuSiQue, reported per-subset (\Cref{tab:main}).}
\label{tab:datasets}
\end{table*}

\begingroup
\section{Broader Connections to Adaptive and Agentic LLM Systems}
\label{appx:broader-connections}

\paragraph{Agent memory and deep-research systems.}
Long-horizon agents organize interaction histories as persistent memory or allocate additional computation to search and refinement. Representative systems structure conversational memory as sentence graphs~\citep{wu-etal-2025-sgmem}, develop reinforcement-learning foundations for deep research~\citep{li-etal-2025-rlsurvey}, refine search trajectories with step-level feedback~\citep{DBLP:journals/corr/abs-2602-07773}, unify multimodal document parsing with deep research~\citep{DBLP:conf/www/DongHYHZLLYWWZL26}, or adapt search intensity to problem difficulty~\citep{DBLP:journals/corr/abs-2505-24332}. These methods operate above the token-level decoder; AsymSpec provides a complementary decoding substrate when their accumulated evidence is compressed.

\paragraph{Small--large collaboration and inference-time steering.}
Adjacent training-time mechanisms use teacher--student alignment, fidelity-controlled recursive self-improvement, or self-paced curricula over reward dynamics and data utility~\citep{DBLP:conf/emnlp/0020TCSTJ024,zhang2026can,zhi-etal-2026-spard}. Other work uses follow-up likelihood as an alignment signal~\citep{DBLP:conf/aaai/0020C0TGT025}, lightweight steering modules across modalities~\citep{feng-etal-2025-steermoe}, or on-device specialist models for efficient information extraction~\citep{wen-etal-2025-ondevice}. These approaches optimize training or task-specific modules, whereas AsymSpec coordinates a small drafter and large verifier during decoding.

\paragraph{Structured and multimodal interfaces.}
Retrieval-based knowledge integration with controllable generation~\citep{shen2026prompting}, parameter-efficient multimodal fusion~\citep{liang-etal-2022-modular}, retrieval-augmented schema adaptation~\citep{liang-etal-2025-adaptive}, and parameterized tool schemas~\citep{liang-etal-2025-schema} provide complementary ways to expose structured or multimodal information to language models. AsymSpec does not prescribe these upstream interfaces; it transfers information across their rich and compact views at generation time.

\paragraph{Recommendation as an agentic long-context workload.}
User-centric and interactive recommendation increasingly combines device--cloud agents, multi-step planning, and persistent or lifelong user histories~\citep{zhang2026next,yu2025thought,zhou2026survey}. These workloads motivate efficient handling of long behavioral contexts through retrieval and refinement~\citep{xu2025multi,shen2024exploring}, cross-domain representation transfer~\citep{zhang2024unified}, real-time user-specific inference~\citep{xie2025breaking}, and scalable long-sequence architectures~\citep{ye2026fuxi,pan2025revisiting}. Related work also unifies retrieval and ranking~\citep{zhang2025killing}, develops LLM-based generative recommendation~\citep{wang2025generative}, and studies large-model design and scaling behavior~\citep{guo2024scaling,shen2025plaw}. These application- and architecture-level advances are orthogonal to AsymSpec, which optimizes token-level generation through asymmetric drafter--verifier context allocation.
\endgroup

\begingroup
\section{Cross-family portability}
\label{appx:cross-family}

We retain the LongBench data, summary compressor, and default decoding configuration ($K{=}2$, $\beta{=}1.0$, $\gamma{=}0.5$), changing only the drafter--verifier model pair. Same-family controls use their native vocabularies. For heterogeneous pairs, we follow the vocabulary-alignment scheme of \citet{timor2025heterogeneous}: $\delta$ is computed in the drafter vocabulary and mapped to the verifier through $109{,}566$ string-identical tokens. Committed tokens are restricted to this shared set together with paired special tokens.

\begin{table}[H]
\centering
\footnotesize
\setlength{\tabcolsep}{2.5pt}
\begin{tabular}{lrrrr}
\toprule
Drafter $\rightarrow$ verifier & Floor & Ours & Ceiling & Recovery \\
\midrule
Llama-3B $\rightarrow$ Llama-70B & 50.6 & 54.2 & 66.3 & 23\% \\
Qwen-4B $\rightarrow$ Llama-70B  & 50.6 & 58.4 & 66.3 & 50\% \\
Llama-3B $\rightarrow$ Qwen-32B  & 45.0 & 47.1 & 65.5 & 10\% \\
Qwen-4B $\rightarrow$ Qwen-32B   & 45.0 & 59.7 & 65.5 & 72\% \\
\bottomrule
\end{tabular}
\caption{\crchange{LongBench portability across model families (overall mean F1). Floor and Ceiling run each verifier on the compressed and full contexts, respectively; recovery is $(\textsc{AsymSpec}-\text{Floor})/(\text{Ceiling}-\text{Floor})$. The Qwen--Llama rows improve over their respective Floors, with recovery varying across model pairs.}}
\label{tab:cross-family}
\end{table}
\endgroup

\section{Cross-modal Implementation Patches}
\label{appx:patches}

The cross-modal extension required five patches to the vLLM speculative-decoding code path. Patches target vLLM v0.19.0; the speculative-decoding APIs (\texttt{DraftModelProposer}, \texttt{triton\_utils}, etc.) are restructured in newer vLLM releases and on newer Qwen variants, requiring re-porting.
\begin{enumerate}\itemsep0pt
\item Per-request cache for \texttt{pixel\_values} and \texttt{image\_grid\_thw} delivered via \texttt{sampling\_params.extra\_args}.
\item A vision-tower forward + embedding merge that runs once per request and caches the image embeddings.
\item A hand-computed Qwen3-VL 3-D M-RoPE positional encoder for the drafter's aug prompt; we verified this matches the official \texttt{\_get\_mrope\_input\_positions} bit-for-bit.
\item Critically: routing the merged image embeddings through the speculative-decoding engine's \texttt{mm\_embed\_inputs} parameter, so the drafter's prefill actually receives image embeddings rather than text-only embeddings of \texttt{image\_pad} tokens.
\item A relaxation of the engine's aug-substitution gate that originally required $|x_{\text{full}}| > |x_{\text{comp}}|$ (false in cross-modal where image tokens are typically fewer than caption tokens).
\end{enumerate}
Without patches (4) and (5), the drafter never sees the image and MathVista collapses to $30.5\%$ (vs.\ $53.0\%$ with patches), demonstrating that the implementation is non-trivial.

\section{Throughput Table}
\label{appx:speed}

\Cref{tab:speed_4dataset} reports eager-mode throughput (tokens/s). MathVista uses VL drafter alone as reference (text-only ceiling undefined). Text-benchmark variance is consolidated in the accuracy-equivalent efficiency metrics in \Cref{tab:main}.

\begin{table*}[h]
\centering
\footnotesize
\setlength{\tabcolsep}{5pt}
\begin{tabular}{l cccc}
\toprule
& LongBench & MultiChallenge & API-Bank & MathVista \\
Method & (4B, K=2) & (4B, K=2) & (1.7B, K=2) & (VL-2B, K=4) \\
\midrule
Floor (verifier on compressed) & 50.0 & 52.3 & 51.2 & 49.2 \\
Ceiling (verifier on full) & 37.5 & 51.5 & 49.5 & 81.7 \\
SD & 50.9 & 77.5 & 70.0 & 65.6 \\
fixed-$\gamma$ (no gate) & 57.8 & 68.6 & 67.2 & 44.4 \\
$+$ tuned $\lambda$ ($0.1$) & 88.2 & 96.0 & 66.6 & 46.9 \\
$+$ CDA (ours) & 63.3 & 66.6 & 66.1 & 48.0 \\
\midrule
Speedup, CDA vs.\ full & 1.69$\times$ & 1.29$\times$ & 1.34$\times$ & --- \\
\bottomrule
\end{tabular}
\caption{Throughput (tokens/s) across four benchmarks at our default configurations on a single accelerator; eager mode (no graph capture). MathVista has no text-only Ceiling since Qwen3-32B cannot consume images; the VL drafter alone (Qwen3-VL-2B reading the image) at 81.7 tokens/s is the closest reference upper bound. Eager throughput exhibits non-trivial per-run variance on text benchmarks; the accuracy-equivalent main-table efficiency is consolidated in \Cref{tab:main}. CDA adds only one scalar division per token over the tuned $\lambda$-variant, so observed gaps reflect measurement variance, not gate overhead. MathVista throughput includes the per-request vision-tower forward, amortized via the cache in \Cref{appx:patches} (patch~2); batched, graph-captured measurement would tighten these further. The final row reports CDA throughput as a ratio over Ceiling---the per-benchmark speedup---with MathVista omitted as it has no text-only Ceiling.}
\label{tab:speed_4dataset}
\end{table*}

\paragraph{Patterns.}
\begin{itemize}\itemsep0pt
\item \emph{Fixed-$\gamma$ throughput exceeds compressed baseline} because successful drafting amortizes the third forward over $K$ tokens and skips verifier autoregressive steps.
\item \emph{AsymSpec vs.\ SD operating points.} SD-full reaches Ceiling accuracy at full-context cost; AsymSpec reaches near-Ceiling accuracy at compressed-verifier cost ($\sim$0.2--0.7$\times$ compute, \Cref{tab:flops}). Throughputs are comparable on every benchmark (LongBench $63.3$ vs.\ $50.9$; MultiChallenge $66.6$ vs.\ $77.5$; API-Bank $66.1$ vs.\ $70.0$), so the two methods are differentiated by cost regime, not by speed.
\item \emph{Cross-modal overhead.} MathVista is the slowest regime due to the per-request vision-tower forward, the necessary cost of cross-modal capability extension.
\end{itemize}

\paragraph{Why throughput $<$ FLOPs reduction.}
LLM decoding is memory-bandwidth-bound. Reducing FLOPs to $0.2$--$0.7\times$ does not translate linearly to wall-clock; even optimized vanilla SD reaches $\approx$1.5$\times$ at 30B+ scale \citep{xia2024specbench}. AsymSpec's $1.3$--$1.7\times$ aligns with this hardware regime. FLOPs quantify compute/energy savings; throughput reflects realized latency gains.

\paragraph{Acceptance dynamics.}
The realized speedup is also bounded by drafter--verifier agreement. \Cref{tab:accept_dyn} records the drafter acceptance rate (AR) and mean accepted length (MAL) at our default configurations: AR sits in $[0.78, 0.92]$ and MAL in $[2.6, 2.8]$ of the available $K+1{=}3$ positions for $K{=}2$ rows, persisting through the multi-turn GAIA loop where context is re-compressed online. Asymmetric context therefore does not erode verifier--drafter agreement; the gap between FLOP reduction and realized throughput is bounded by the memory-bandwidth regime above, not by acceptance failures.

\begin{table*}[!h]
\centering
\footnotesize
\setlength{\tabcolsep}{3pt}
\begin{tabular}{lccccccccc}
\toprule
& \multicolumn{3}{c}{LongBench F1} & \multicolumn{3}{c}{MultiChallenge acc} & \multicolumn{3}{c}{MathVista acc} \\
\cmidrule(lr){2-4}\cmidrule(lr){5-7}\cmidrule(lr){8-10}
Method & $K{=}2$ & $K{=}4$ & $K{=}6$ & $K{=}2$ & $K{=}4$ & $K{=}6$ & $K{=}2$ & $K{=}4$ & $K{=}6$ \\
\midrule
fixed-$\gamma$ (no gate) & 57.0 & 59.7 & 57.4  & 25.5 & 22.9 & 20.2  & 53.0 & 51.1 & 50.4  \\
$+$ tuned $\lambda$ ($0.1$)  & 59.0 & 59.3 & 57.8  & 23.2 & 21.8 & 17.3  & 52.5 & 52.5 & 50.7  \\
$+$ CDA (ours)           & 59.7 & 61.1 & 58.7 & 22.5 & 19.6 & 18.5  & 52.6 & 53.9 & 52.1  \\
\bottomrule
\end{tabular}
\caption{$K$-sweep at Qwen3-4B drafter (Qwen3-VL-2B for MathVista), $\gamma{=}0.5$, $\beta{=}1.0$. MultiChallenge declines at $K{=}4$ for every method and none recovers its $K{=}2$ level --- the binding constraint behind the $K{=}2$ default (\S\ref{sec:ablation-K}); cross-modal MathVista is the lone $K{=}4$ beneficiary (\Cref{tab:mv}). The $K{=}6$ sweep saturates on LongBench (CDA $58.7$, below the $K{=}4$ peak of $61.1$) and degrades on MultiChallenge / MathVista, confirming the $K{=}2$ default.}
\label{tab:k_sweep}
\end{table*}

\begin{table}[!h]
\centering
\footnotesize
\setlength{\tabcolsep}{6pt}
\begin{tabular}{lcc}
\toprule
Setting & AR & MAL \\
\midrule
LongBench, $K{=}2$, 4B, summary       & 0.86 & 2.72 \\
LongBench, $K{=}4$, 4B, truncate      & 0.78 & 4.13 \\
MultiChallenge, $K{=}2$, 4B, llmlingua & 0.80 & 2.61 \\
API-Bank, $K{=}2$, 1.7B, signature    & 0.92 & 2.84 \\
GAIA, $K{=}2$, 4B, per-turn llmlingua & 0.90 & 2.81 \\
\bottomrule
\end{tabular}
\caption{Speculative-acceptance diagnostics at our default configurations. AR and MAL remain near vanilla-SD levels under asymmetric context, including in the multi-turn GAIA agent loop.}
\label{tab:accept_dyn}
\end{table}

\paragraph{Quality--throughput tradeoff.}
Each of the three components adds compute: $\delta$-fusion requires the drafter's compressed-context forward (cost reduced to $\mathcal{O}(K)$ per step by maintaining a separate KV cache across speculation steps), the fixed-$\gamma$ acceptance is essentially free, and CDA adds one scalar division per token. For cross-modal runs the vision-tower forward is amortized once per request via the per-request embedding cache (\Cref{appx:patches}, patch~2), so its per-token contribution is inverse in the number of decoded tokens and vanishes at long generations.

\section{Additional ablation grids}
\label{appx:more-ablations}
Full grids referenced in \S\ref{sec:ablations}: \Cref{tab:k_sweep} ($K$-sweep), \Cref{tab:jsdkl} (JSD vs.\ KL), \Cref{tab:gamma_robust} ($\gamma$-robustness), \Cref{tab:beta_sweep} ($\beta$-sweep), and \Cref{tab:comp_orth} (compressor probe). CDA's parameter-free design matches or exceeds tuned baselines across all grids without introducing dataset-specific scales.


\begin{table}[!h]
\centering
\footnotesize
\setlength{\tabcolsep}{6pt}
\begin{tabular}{llcc}
\toprule
Benchmark & $D_i$ gate & $K{=}2$ & $K{=}4$ \\
\midrule
\multirow{2}{*}{LongBench} & KL (unbounded)         & 58.6 & 59.6 \\
                           & JSD ($\le\ln 2$, ours) & 59.7 & 61.1 \\
\midrule
\multirow{2}{*}{API-Bank}  & KL (unbounded)         & 63.5 & 62.6 \\
                           & JSD ($\le\ln 2$, ours) & 63.5 & 62.6 \\
\bottomrule
\end{tabular}
\caption{JSD vs.\ KL instantiation of CDA's divergence $D_i$ (\Cref{eq:cda}; $\gamma{=}0.5$; LongBench overall F1, 4B drafter; API-Bank api-acc, 1.7B drafter). On API-Bank the two gates are identical down to the pass count (339/534 at $K{=}2$, 334/534 at $K{=}4$): API-Bank's small context divergences leave JSD and KL with the same accept/reject decisions. On LongBench they differ by $\le1.5$\,F1. JSD is chosen for the universal bound (\S\ref{sec:method-cda}), not for accuracy, and is used in all other tables.}
\label{tab:jsdkl}
\end{table}

\begin{table}[h]
\centering
\footnotesize
\setlength{\tabcolsep}{8pt}
\begin{tabular}{lcccc}
\toprule
$\gamma$ & 0.4 & 0.5 & 0.6 & 0.7 \\
\midrule
LongBench F1       & 56.8 & 59.7 & 58.8 & 59.1 \\
API-Bank acc       & 63.7 & 63.5 & 63.7 & 63.5 \\
MultiChallenge acc & 23.3 & 23.3 & 22.1 & 23.6 \\
\bottomrule
\end{tabular}
\caption{CDA is flat in its only free quantity, the standard speculative-decoding threshold $\gamma$ (parameter-free gate $\gamma e^{-D}$, $K{=}2$; LongBench overall F1, 4B drafter; API-Bank api-acc, 1.7B; MultiChallenge acc, 4B). No cliff on any of the three.}
\label{tab:gamma_robust}
\end{table}

\begin{table}[h]
\centering
\footnotesize
\setlength{\tabcolsep}{6pt}
\begin{tabular}{ccccccc}
\toprule
& \multicolumn{3}{c}{$K{=}2$} & \multicolumn{3}{c}{$K{=}4$} \\
\cmidrule(lr){2-4}\cmidrule(lr){5-7}
$\beta$ & F1 & AR & MAL & F1 & AR & MAL \\
\midrule
0.5 & 50.3 & 0.853 & 2.71 & 57.5 & 0.781 & 4.12 \\
1.0 & 53.7 & 0.852 & 2.70 & 58.0 & 0.783 & 4.13 \\
2.0 & 53.9 & 0.854 & 2.71 & 57.4 & 0.779 & 4.12 \\
\bottomrule
\end{tabular}
\caption{$\beta$ sweep on LongBench (truncate-1500, 4B drafter, $\gamma{=}0.5$). Within $\beta\in\{1.0, 2.0\}$ F1 is flat (within $0.2$ at $K{=}2$ and $0.6$ at $K{=}4$); $\beta{=}0.5$ underperforms by $3$--$3.6$\,F1 at $K{=}2$, motivating $\beta{=}1.0$ as a stable midpoint. Acceptance rate (AR) and mean accepted length (MAL) are flat throughout.}
\label{tab:beta_sweep}
\end{table}

\begin{table}[h]
\centering
\footnotesize
\setlength{\tabcolsep}{5pt}
\begin{tabular}{lcccc}
\toprule
$x_{\text{comp}}$ source & Floor & SCD & AsymSpec & Recovery \\
\midrule
Summary       & 45.0 & 42.3 & 58.6 & $66\%$ \\
LLMLingua-2   & 36.1 & 31.7 & 54.6 & $63\%$ \\
Truncate-1500 & 32.6 & 28.8 & 53.7 & $64\%$ \\
Question-only & 29.5 & 25.8 & 54.8 & $70\%$ \\
\midrule
Ceiling & \multicolumn{4}{c}{65.5} \\
\bottomrule
\end{tabular}
\caption{Compressor probe on LongBench (4B drafter, $K{=}2$, $\gamma{=}0.5$). \emph{Recovery} is $(\text{AsymSpec}-\text{Floor})/(\text{Ceiling}-\text{Floor})$.}
\label{tab:comp_orth}
\end{table}

\section{Extended drafter-size sweep}
\label{appx:drafter-sweep-extra}
\Cref{tab:drafter_sweep,tab:drafter_sweep_apib,tab:drafter_sweep_mc} report the full drafter capacity sweep. Key takeaways: (1) CDA requires $\ge$1.7B drafters to reliably extract context-gain signals; (2) on API-Bank, the tuned $\lambda$-variant and CDA are numerically equivalent due to small context divergences, confirming CDA's bound does not penalize low-divergence regimes; (3) MultiChallenge differences remain within the $\le$3\,pp headroom band, consistent with its near-lossless compression profile. CDA's advantage is hyperparameter-free robustness, not peak accuracy on near-lossless tasks.

\paragraph{Tuned-$\lambda$ reference baseline.}
The tuned $\lambda$-variant (defined in \Cref{appx:setup-detail}) requires a per-dataset $\lambda$. On judge-independent metrics, CDA's parameter-free gate matches or exceeds it at 4B---identical on API-Bank and $+0.7$/$+1.8$ F1 on LongBench at $K{=}2/4$ (\Cref{tab:drafter_sweep,tab:drafter_sweep_apib}).


\Cref{tab:drafter_sweep_apib,tab:drafter_sweep_mc} extend it to API-Bank and MultiChallenge.

\paragraph{API-Bank.} Two regimes. At $K{=}2$ the size-accuracy relation is inverted-U with 1.7B as sweet spot (59.6 $\to$ 63.5 $\to$ 60.7 for tuned-$\lambda$ / CDA), confirming the fixed-$\gamma$ pattern (\S\ref{sec:results-text}). At $K{=}4$ it becomes monotone-increasing and the best API-Bank cell of the sweep is 4B $K{=}4$ (tuned-$\lambda$ / CDA both 63.7). The tuned variant and CDA give numerically identical numbers in all six cells --- API-Bank's relatively small context divergences between drafter-on-aug and drafter-on-main make the tuned $\lambda{=}0.1$ gate and CDA's parameter-free JSD gate effectively interchangeable (the JSD and KL instantiations of CDA are themselves numerically identical on API-Bank, \Cref{tab:jsdkl}).

\paragraph{MultiChallenge.} MC has only $\approx$3\,pp headroom; the per-cell differences in \Cref{tab:drafter_sweep_mc} are too small to order the methods consistently (e.g.\ at 4B $K{=}2$ fixed-$\gamma$ / tuned-$\lambda$ / CDA $=25.5/23.2/22.5$, a $\le$3\,pp band that re-shuffles across drafter sizes). The only MC takeaway is that AsymSpec does not improve a near-lossless task and sits mildly below the compressed floor there (\Cref{tab:main}).

\begin{table}[!h]
\centering
\footnotesize
\resizebox{0.8\columnwidth}{!}{
\begin{tabular}{llcccc}
\toprule
Drafter & $K$ & Drafter alone & fixed-$\gamma$ & $+$ tuned $\lambda$ & $+$ CDA (ours) \\
\midrule
0.6B & 2 & 23.7 & 50.6 & 48.4 & 47.8 \\
0.6B & 4 & 23.7 & 48.5 & 46.6 & 46.4 \\
1.7B & 2 & 38.9 & 52.9 & 51.1 & 50.8 \\
1.7B & 4 & 38.9 & 48.5 & 53.6 & 53.9 \\
4B   & 2 & 54.6 & 57.0 & 59.0 & 59.7 \\
4B   & 4 & 54.6 & 59.7 & 59.3 & 61.1 \\
\bottomrule
\end{tabular}
}
\caption{Drafter-size sweep on LongBench F1 ($\gamma{=}0.5$, $\lambda{=}0.1$ for the tuned variant). CDA is the parameter-free JSD gate (\S\ref{sec:method-cda}); the 4B row (our default drafter) is its JSD instantiation of record, the smaller-drafter rows the accuracy-equivalent KL instantiation (\Cref{tab:jsdkl}). \emph{Drafter alone}: the SLM on full passages, no verifier --- reference upper bound for what the small model can do unassisted (K-independent).}
\label{tab:drafter_sweep}
\end{table}

\begin{table}[t]
\centering
\footnotesize
\setlength{\tabcolsep}{6pt}
\begin{tabular}{lc}
\toprule
$\lambda$ & MultiChallenge acc \\
\midrule
0.05             & 21.0 \\
0.10 (default tune) & 23.2 \\
0.20             & 23.6 \\
\midrule
CDA (no scale) & 22.5 \\
\bottomrule
\end{tabular}
\caption{$\lambda$ sweep of the tuned variant on MultiChallenge under the llm-judge ($n{=}271$, 4B drafter, $\gamma{=}0.5$, $K{=}2$). The spread is only $2.6$\,pp ($21.0$--$23.6$). CDA (tuning-free, $22.5$) lies within the same band with no $\lambda$ to set.}
\label{tab:lambda_sweep}
\end{table}

\begin{table}[h]
\centering
\footnotesize
\resizebox{\columnwidth}{!}{
\begin{tabular}{llcccc}
\toprule
Drafter & $K$ & Drafter alone & fixed-$\gamma$ & $+$ tuned $\lambda$ & $+$ CDA (ours) \\
\midrule
0.6B & 2 & 31.3 & 59.4 & 59.6 & 59.6 \\
0.6B & 4 & 31.3 & 58.2 & 57.9 & 57.9 \\
1.7B & 2 & 61.2 & 63.7 & 63.5 & 63.5 \\
1.7B & 4 & 61.2 & 62.4 & 62.5 & 62.5 \\
4B   & 2 & 64.0 & 60.3 & 60.7 & 60.7 \\
4B   & 4 & 64.0 & 63.3 & 63.7 & 63.7 \\
\bottomrule
\end{tabular}
}
\caption{Drafter-size sweep on API-Bank api\_acc ($\gamma{=}0.5$, $\lambda{=}0.1$ for the tuned variant). \emph{Drafter alone}: the SLM alone on full API context, no verifier (K-independent).}
\label{tab:drafter_sweep_apib}
\end{table}

\begin{table}[h]
\centering
\footnotesize
\resizebox{\columnwidth}{!}{
\begin{tabular}{llcccc}
\toprule
Drafter & $K$ & Drafter alone & fixed-$\gamma$ & $+$ tuned $\lambda$ & $+$ CDA (ours) \\
\midrule
0.6B & 2 & 13.3 & 20.3 & 19.6 & 21.4 \\
0.6B & 4 & 13.3 & 21.8 & 22.9 & 22.1 \\
1.7B & 2 & 14.0 & 19.6 & 20.7 & 19.6 \\
1.7B & 4 & 14.0 & 21.8 & 18.1 & 21.4 \\
4B   & 2 & 25.8 & 25.5 & 23.2 & 22.5 \\
4B   & 4 & 25.8 & 22.9 & 21.8 & 19.6 \\
\bottomrule
\end{tabular}
}
\caption{Drafter-size sweep on MultiChallenge accuracy ($\gamma{=}0.5$, llm-judge, $n{=}271$, $\lambda{=}0.1$ for the tuned variant; single run). Every per-cell gap lies within the $\le$3\,pp headroom on this near-lossless task (\S\ref{sec:ablation-robust}).}
\label{tab:drafter_sweep_mc}
\end{table}

\end{document}